# Industrial object, machine part and defect recognition towards fully automated industrial monitoring employing deep learning. The case of multilevel VGG19

Ioannis D. Apostolopoulos [1]*, Mpesiana Tzani[2]

1. Department of Medical Physics, School of Medicine, University of Patras, GR 265-00 Patras;
   ece7216@upnet.gr
2. Department of Electrical and Computer Engineering, University of Patras, GR 265-00

* Correspondence: Ioannis D. Apostolopou1los

**Abstract:**

Modern industry requires modern solutions for monitoring the automatic production of goods. Smart monitoring of the functionality of the mechanical parts of technology systems or machines is mandatory for a fully automatic production process. Although Deep Learning has been advancing, allowing for real-time object detection and other tasks, little has been investigated about the effectiveness of specially designed Convolutional Neural Networks for defect detection and industrial object recognition. In the particular study, we employed six publically available industrial-related datasets containing defect materials and industrial tools or engine parts, aiming to develop a specialized model for pattern recognition. Motivated by the recent success of the Virtual Geometry Group (VGG) network, we propose a modified version of it, called Multipath VGG19, which allows for more local and global feature extraction, while the extra features are fused via concatenation. The experiments verified the effectiveness of MVGG19 over the traditional VGG19. Specifically, top classification performance was achieved in five of the six image datasets, while the average classification improvement was 6.95%.

**1. Introduction**

Modern production automation systems have opened up great horizons, simplifying many functions of the production process, accelerating the production, repair, and transportation of products. The introduction of several automation systems requires the corresponding automatic control of the systems for the timely and valid detection of errors, the confrontation of dangerous situations, and the smooth maintenance of the machines (**Diez-Olivan 2019**). Such procedures are required to be carried out in real-time with the utilization of appropriate equipment. Hence, humanity aims to convert the production process into a smart one.

In recent years, object detection on images has been one of the most actively researched Artificial Intelligence (AI) tasks (**Han 2018**). It refers to technologies that, through specific algorithms based on Machine Learning, can classify specifically targeted subjects. Computer vision, which we describe as the art and science of making computers recognize digital images (**Khan 2018**), is strongly connected. Accordingly, object recognition is a computer vision technique used to recognize and find objects within an image or video. Specifically, object detection draws bounding boxes around these detected objects to identify where the objects are in (or how they pass through) a particular scene. Object recognition consists of recognizing, identifying, and locating objects within an image with a given amount of confidence. This method involves the work of classification, tagging, detection, and segmentation.

Object recognition is now present in various applications such as scenario comprehension, video surveillance, robotics, and motor vehicles. These are the applications that have ignited significant research in the field of computer vision over the last decade. Visual recognition systems, including image detection or identification, have achieved tremendous research potential due to considerable growth in neural networks, especially deep Learning, and have achieved remarkable efficiency.

In several industry sectors, object recognition has many exciting uses. For example, automated systems can be built by object recognition to recognize defective parts or tools for immediate replacement and also to detect individual parts that require repair and replacement during the manufacturing process. Even the ability to scan for objects and measure their number in images is essential for a company, particularly in the industry. We are talking about a large number of machinery. The regular tasks of manually measuring the number of different components or objects are an essential part of the working time of the specialists. The application of research in the field of deep convolutional neural networks to the tasks of detection and classification can also help to automate specific repetitive tasks.

Recently, object recognition from digital images or videos made significant progress, thanks to the development of new image processing techniques, also known as Deep Learning with Convolutional Neural Networks. Deep Learning alludes to various Machine Learning methods and has already succeeded in speech recognition, image recognition, natural language processing, and more. Convolutional Neural Networks are a special type of traditional Neural Networks, which employ the convolution process to analyze the input data distributions and generate potentially powerful features related to a specific domain.

Deep Learning could be extremely useful in industrial projects. In recent years, the research community has put particular emphasis on developing such image processing systems for materials in manufacturing as a way of promoting all sorts of functions that make production time-consuming and expensive.

Caggiano et al. (**Caggiano 2019**) developed a machine learning method based on Deep Convolutional Neural Network (DCNN) to detect defects based on SLM non-compliance through automated image processing. In particular, a machine learning method has been developed for the online detection of defects through automated image processing, leading to the timely detection of defective sections of a material due to non-compliance with the Selective Laser Melting (SLM) metal powder process. During layer-by-layer SLM processing, the images in this study were obtained and the analysis was performed using a Deep Convolutional Neural Network model based on two currents and automatic image learning and feature fusion achieved the identification of the defective condition-related SLM pattern. The efficacy of DCNNs in various scenarios of manufacturing (inspection, motion detection, and more) has been demonstrated in recent years (**Gu 2017, Wang 2018, Wang 2018b**). The work of Fu et al. (**Fu 2019**), which focuses on automated visual identification of steel surface defects, is impressive and can make a major contribution to functionality in order to facilitate quality control of the output of steel strips. In order to achieve fast and accurate classification of defective steel surfaces in their work, they present an effective model of a convolutional neural network (CNN), which emphasizes the training of low-level features and incorporates multiple receptive fields. Their methodology focuses on three basic modules, first on the use as a fundamental architecture of pre-trained SqueezeNet (**Iandola 2016**). Second, in a series of reinforced diversity surface steel defect data containing extreme non-uniform lighting, camera noise and motion blur, the use of only a limited amount of defect-specific training samples for high-accuracy detection. Finally, by running over 100 fps on a machine equipped with a single NVIDIA TITAN X graphics processing unit (12G memory), the lightweight CNN model they used will fulfill the requirement for real-time online inspection. Another approach from Wang et al (**Wang 2019**). refers to a new mechanical vision inspection system focused on learning to identify and classify a faulty product without loss of precision. The Gauss filter is used explicitly in this work where it is executed in the obtained image for the first time to decrease unintended noise. The contribution of the work lies in the unloading of the computational burden of the next identification process because of the export of a region of interest (ROI) based on the transformation of Hough to eliminate the irrelevant context. In order to achieve a good balance between recognition accuracy and computational performance, the construction of the recognition unit is based on a convergent neural network, while the remaining inverted block is implemented as the basic block. By using the proposed approach with a large number of data sets consisting of inaccurate and defective bottle images, superior control efficiency is achieved. They emphasize that the monitoring system used in this work is capable of covering both precision and effectiveness by combining traditional methods of image processing and a light deep neural network.

Despite the remarkable progress and the numerous deep learning proposals for object and defect recognition, little has been said about the effectiveness of universal models, specifically designed for object recognition in images with industrial content, such as images illustrating machines, metallic parts, tools, etc. Motivated by the recent success of the deep learning model called VGG19 (Virtual Geometry Group) (**Simonyan 2014**), we propose a universal VGG-based deep learning framework capable of recognizing multiple defects and objects from multiple image sources. To achieve this, we based our experiments on two strategies as follows:

a) The classic architecture of VGG19 was modified to allow for more feature extraction coming from the early convolutional layers, where low-level features are extracted.
b) Rather than evaluating it solely on one image dataset, the proposed model was tested in several industrial-related image datasets to investigate its general efficiency.

## 2. Material and Methods

### *2.1 Deep Learning with Multilevel Virtual Geometry Group 19 network (MVGG19)*

The main advantage of deep Learning with Convolutional Neural Networks lies in extracting new features from the input data distributions (i.e., images), thereby bypassing the manual feature extraction process, which is traditionally performed in image analysis tasks with machine learning methods (**Lin 2017**).

Each convolution layer in a CNN is processing the previous layer's output by applying new filters and extracting new features (**LeCun 2015**). Since the convolutional layers are stacked together, a hierarchical process takes place. In essence, features directly from the original image are only extracted by the first convolutional layer, whereas the other layers process each other's outputs. In this way, a slow introduction to large amounts of filters is achieved, while underlying features may be revealed during the last layers. The general rule of thumb relates the network's effectiveness with the number of convolutional layers (**Shin 2016**). This is why deep networks are generally superior, provided that an adequate amount of image data is present. In cases where the dataset's size is not large enough to feed a deep network, three solutions are commonly proposed:
a) The selection of a simpler CNN, which contains less trainable parameters and fits in the particular data well
b) Transfer learning (**Kornblith 2019**), utilizing deep and complex CNNs but freezing their layers, thereby decreasing the trainable parameters and allowing for knowledge transfer, following their training on large image datasets.
c) Data augmentation (**Wong 2016**) methods to increase the training set size.

In the particular study, although the size of the training data is not negligible, we propose both transfer learning and data augmentation to increase the training set further and train a robust model with the ability to generalize.

Because the low-level features, which are generally extracted from the first (or bottom) layers of the CNN, may be useful for the classification task (**Yue-Hei Ng 2015**), a completely hierarchical network may lack on this front. Typical CNNs fuse the initially extracted features with deeper features because one convolution comes after the other. To circumvent this setback, we investigate and experiment with a modification of the original structure of the CNN to allow for parallel feature extraction, which is achieved utilizing many paths.

Inspired by the Victual Geometry Group network (**Simonyan 2014**), in this study, a novel modification is proposed and analyzed. Multipath VGG19, as shown in **Figure 1**, takes full advantage of the traditional hierarchical structure of VGG19 while it connects early and late convolutional blocks together by constructing extra feature map processing paths. After the second, third, and fourth max-pooling layers, the extracted feature maps are processed by "BD" layers, which apply batch normalization, dropout, and global average pooling. The three BD outputs and the sequential output are concatenated and fed to a neural network of 2500 neurons. A softmax classifier performs the feature classification according to the desired number of classes (in **Error! Reference source not found.**, two arbitrary classes are shown). The source code for constructing MVGG19 from scratch is provided.

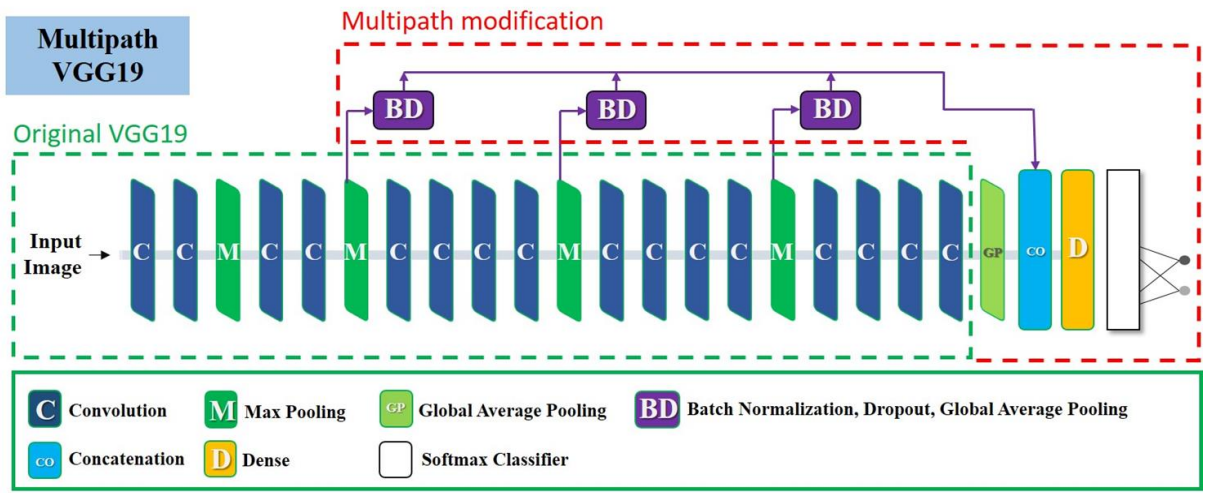

**Figure 1.** Multipath VGG19

MVGG19 is fine-tuned (**Tajbakhsh 2016**) by maintaining the top (i.e., the last) layer trainable, while the rest of the layers are frozen to retain their weights obtained from its initial training. We used batch normalization and 50% dropout after the dense layer of 2500 neurons. Optimization was achieved employing the default parameters of the Adam (**Kingma 2014**) optimization algorithm. All experiments were performed in a python environment making use of the Keras library. An Intel Core i5-9400F CPU at 2.90GHz computer equipped with 6Gb RAM and a GeForce RTX 2060 Super was the primary infrastructure for the experiments.

## 2.2 Image datasets for industrial object recognition and defect detection

For the particular study, we selected six publicly available sets of images related to industrial applications. The characteristics of the evaluation datasets are described in this section.

**Industrial Dataset of Casting Production (Casting dataset)**

This dataset is of casting manufacturing product. Casting is a manufacturing process in which a liquid material is usually poured into a mould, which contains a hollow cavity of the desired shape, and then allowed to solidify. All included images are top views of the submersible pump impeller. The total images are 7348, while their size is 300x300 pixels. Two categories are describing the contents of each image, namely defective object and normal. The images were very well organized, and no image preprocessing was required. The dataset is openly available in Kaggle (**Dabhi 2020**). Examples of the two classes are depicted in **Figure 2**. The task of the proposed MVGG19 model is to correctly detect the problematic parts.

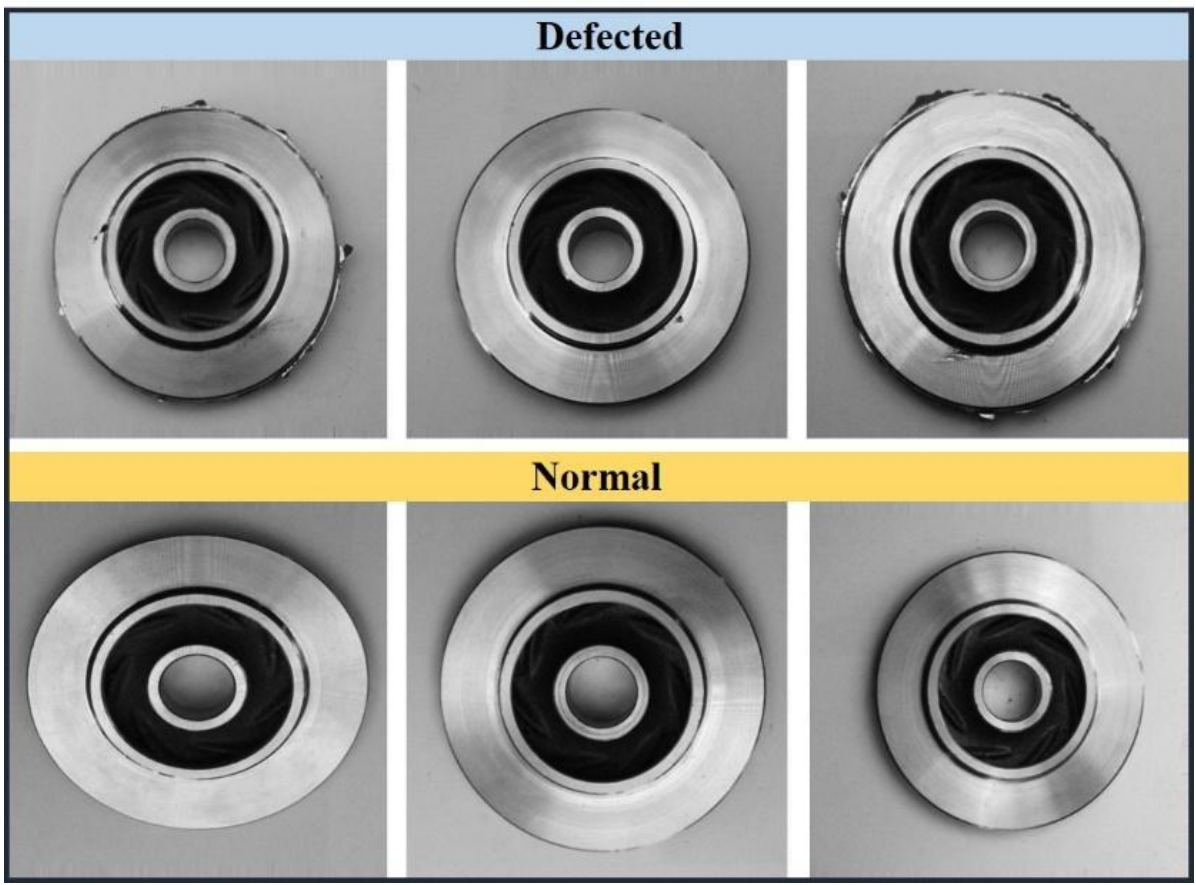

**Figure 2**. Industrial Dataset of Casting Production dataset

**Defects location for metal surface dataset (Defect dataset)**

Known initially as GC10-DET dataset (**Lv 2020**), this set refers to ten types of surface defects, i.e., punching (Pu), weld line (Wl), crescent gap (Cg), water spot (Ws), oil spot (Os), silk spot (Ss), inclusion (In), rolled pit (Rp), crease (Cr), waist folding (Wf). The collected defects are on the surface of the steel sheet. The dataset includes 3570 gray-scale images. Distinguishing between those types of defection is crucial for the industry. It can significantly contribute to the prevention of malfunction, the real-time identification of defects on a variety of essential materials found in factories, processing plants, and logistics. The dataset is publicly available at Github. **Figure 3** illustrates some examples.

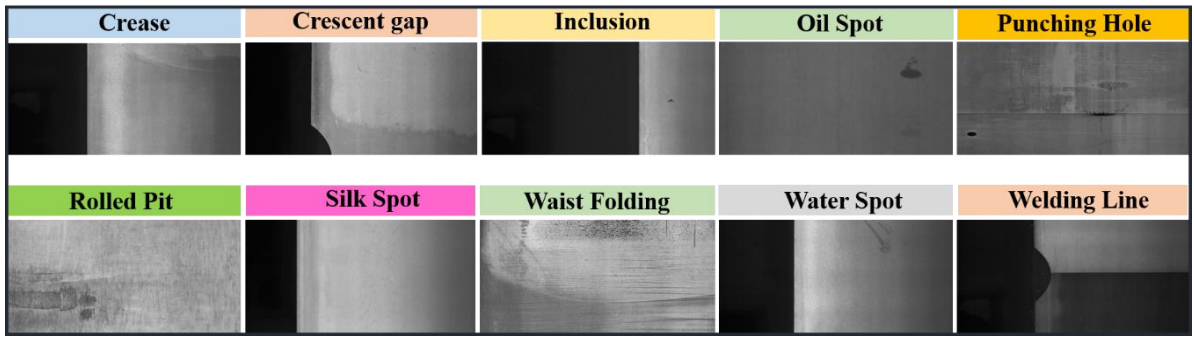

**Figure 3.** Defects location for metal surface dataset

**Magnetic Tile Defect dataset (Magnetic dataset)**

The present dataset (**Huang 2018**) is available for research purposes in a variety of repositories. The images of 6 common magnetic tile defects were collected, while the dataset contains annotations for segmentation tasks. The dataset contains 1243 images of those 6 classes. Samples are illustrated in **Figure 4**.

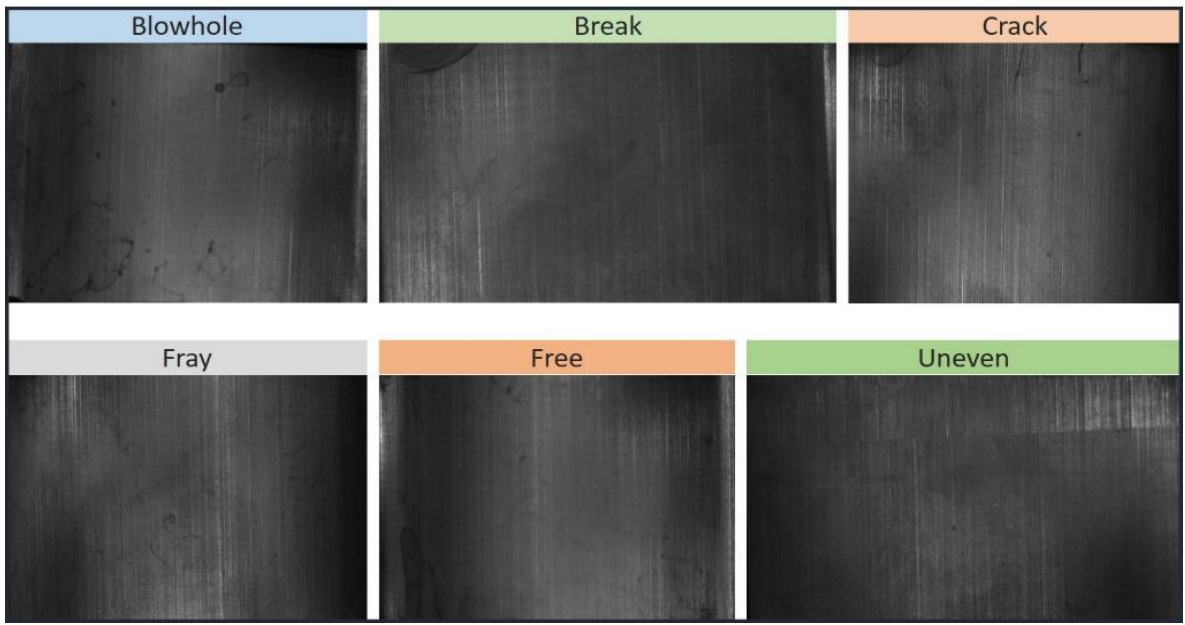

**Figure 4.** Magnetic Tile Defect dataset

**Object Recognition in Industry dataset (Tech dataset)**

Focused on industrial applications, The MVTec Industrial 3D Object Detection Dataset (MVTec ITODD) is a public dataset object detection and pose estimation in 2D or 3D (**Drost 2017).** It consists of 28 object classes and more than 3500 labeled images of those objects. For the particular task, 10 object types are selected, namely cylinder, planar bracket, star, fuse, box, round engine cooler, cap, engine cover, car rim, and bracket screw. Samples of those objects are illustrated in **Figure 5**.

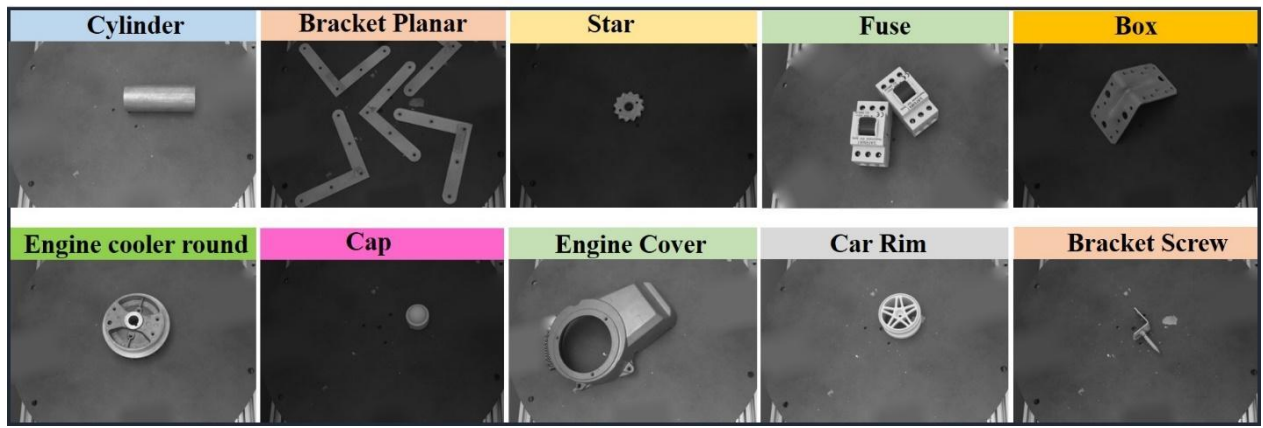

**Figure 5.** Object Recognition in Industry dataset

**Bridge Crack recognition dataset (Bridge dataset)**

This dataset's original name is SDNET2018 (**Maguire 2018**), which is a publicly available annotated dataset intended for the evaluation of artificial intelligence algorithms. It contains over 56 thousand images of cracked and non-cracked concrete bridge decks, walls, and pavements, where cracks are as narrow as 0.06 mm and as wide as 25 mm. Shadows, surface roughness, scaling, edges, holes, and background debris are artifacts included in the images, making the recognition task more challenging. Two classes are generated from this dataset, namely "cracked" and "ok." **Figure 6** illustrates some samples of the two classes.

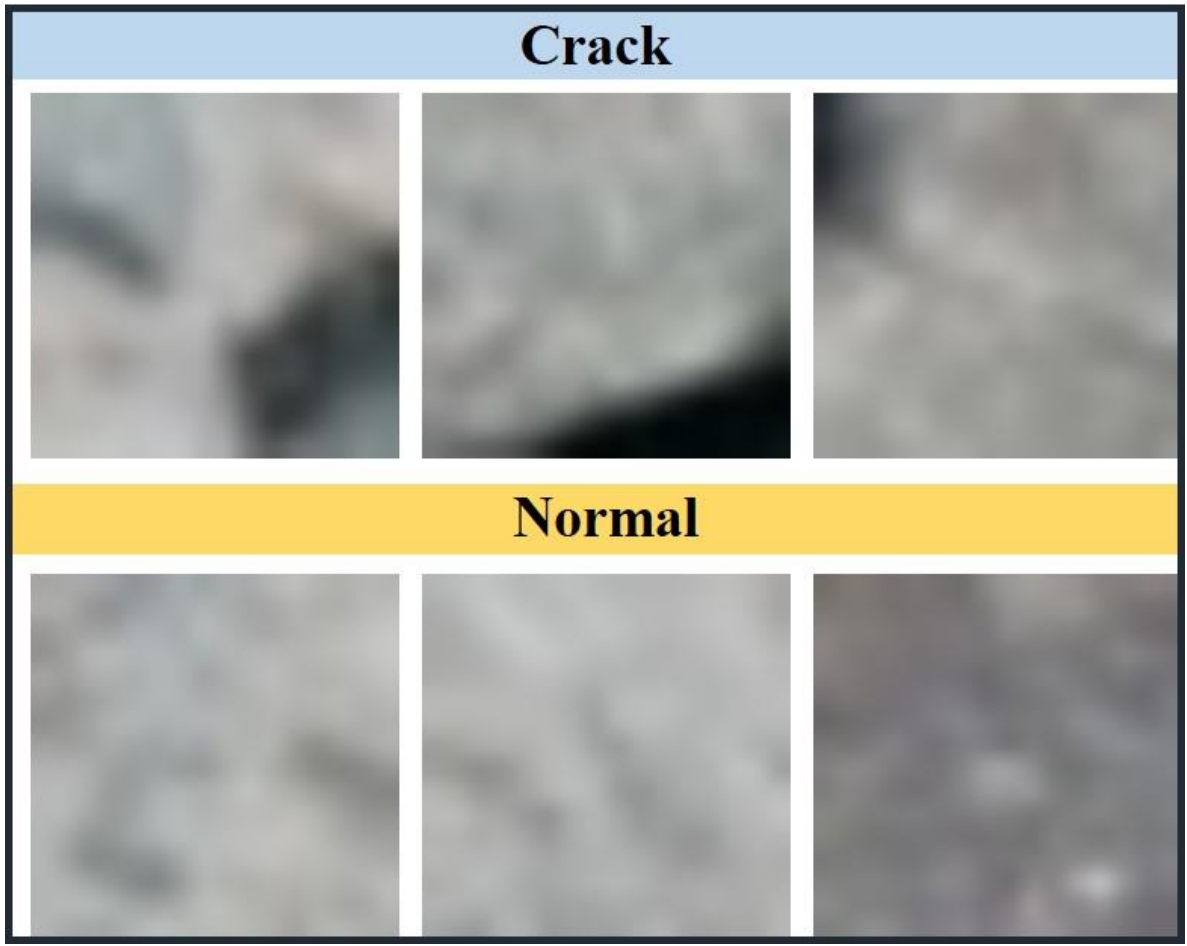

**Figure 6**. Bridge Crack recognition dataset

**Solar Cell defect probability dataset (Solar dataset)**

Solar Cell defect probability dataset was acquired from Github (**Buerhop-Lutz 2018, Deitsch 2019, Deitsch 2020**). It contains functional and defective solar cell surfaces with a variety of degradation degrees. For the particular experiment, two classes are created, namely "defect" and "ok," where images annotated with a degree above zero are considered defective. As mentioned in the repository, all the included images had been normalized with respect to their size and their perspective. Prior to solar cell extraction, the distortions induced by the camera utilized has been eliminated. The overall size of the dataset is 2624 images, while each image size is 300x300 pixels of 8-bit. 44 different solar modules were examined. In **Figure 7**, a dataset sample is illustrated.

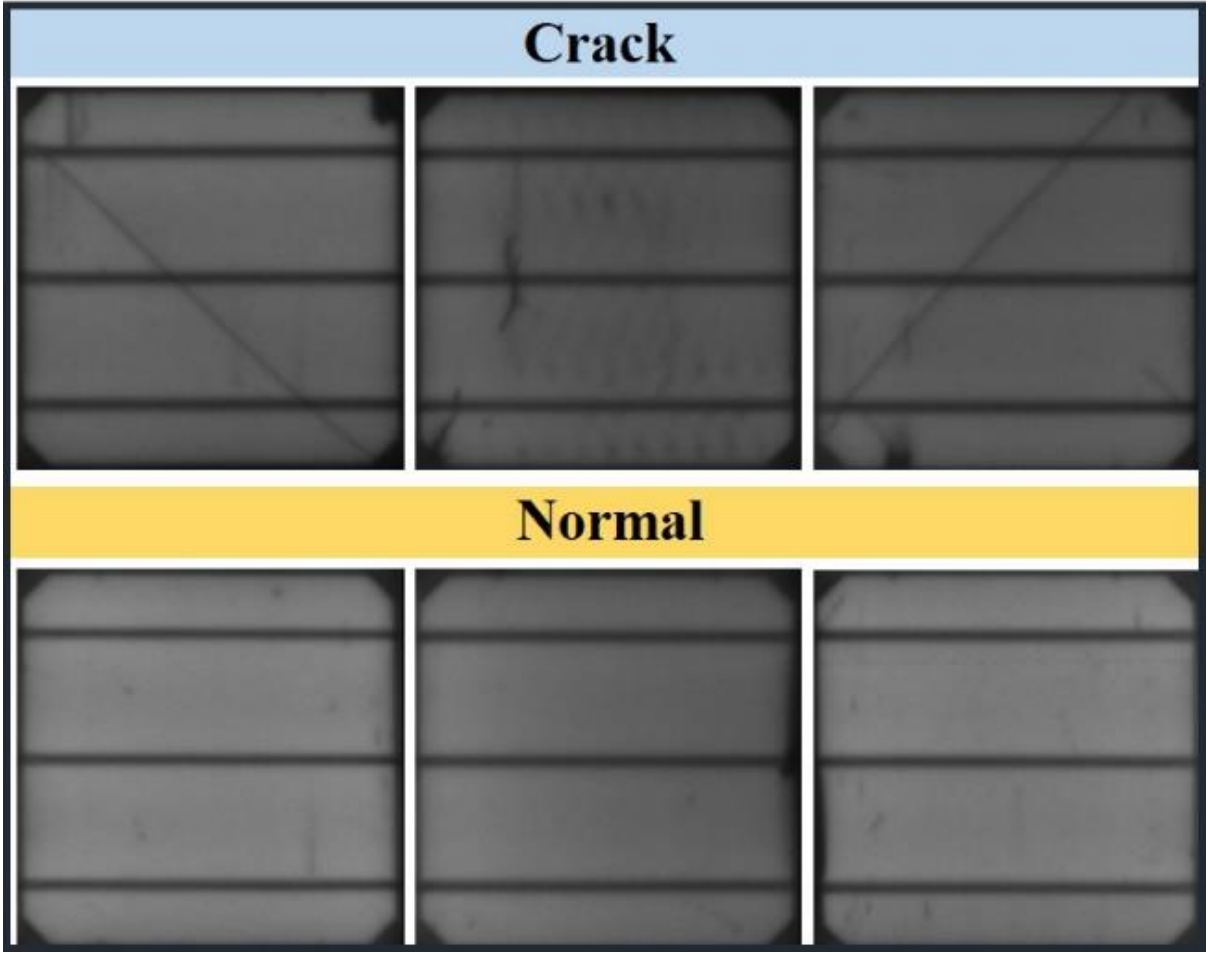

**Figure 7.** Solar Cell defect probability dataset

*2.3 Data augmentation techniques*

Data augmentation is an essential step in deep learning applications and research, mainly utilized for two reasons (**Fawzi 2016**). The first reason is the data scarcity, which impedes deep learning models adapt to the domain of interest and develop their learning capabilities. Few images are commonly not enough for a deep learning framework to train on. Especially in cases where the classification should be based on deep features and not obvious and low-level characteristics (e.g., colors). With data augmentation, the initial training set can be broadly expanded by applying various transformations to the original images. In this way, the model learns to ignore useless characteristics and improves its spatial capabilities. For example, applying random rotations directs the model towards seeking patterns in moving positions and not stable.

By data scarcity, not only is the shortage of data considered, but also the shortage of data covering the diversity of the statistical population of the target classification. For example, a dataset may contain thousands of images illustrating the front of a car, based upon which the developed model shall distinguish between old-fashioned and modern cars. Supplying a new, unseen image to the model for prediction, where the rear of a random car is depicted, may lead the model to the wrong conclusion since the initial training had been made utilizing solely front-view images. Data augmentation may modify or generate completely new images, capturing essential features found in both front-view and rear-view images, thereby expanding the training sets' diversity. This is the second reason why data augmentation is preferable.

In the present research, we applied the following augmentations to the training sets to expand the available data and to increase the generalization capabilities of the experimental deep learning networks:
- a. Random Gaussian Noise
- b. Random Rotations
- c. Horizontal and Vertical Flips
- d. Height and Width shifts

**3. Results**

The epochs of training and the batch size were adjusted for each dataset to perform an optimal training fitting to the particular computational infrastructure. For the evaluation of the performance, 10 fold cross-validation was preferred. During this process, ten independent training-testing phases took place. For each phase, a different part of the dataset was selected to serve as a test (i.e., hidden) set, while the rest of the dataset was utilized for training. The results are aggregated, and the performance metrics correspond to the mean of the metrics recorded during each phase.

For the majority of the datasets, the proposed network achieves top accuracy and Area Under Curve score. Specifically, the best classification accuracy is obtained for the Bridge dataset, with 99.02%. The second-best accuracy obtained is 97.88% in the Defect dataset. The rest of the accuracy results are 94.23% for the Tech dataset, 92.67% for the Magnetic dataset, 77.62% for the Casting dataset, and 76.68% for the Solar dataset. **Table 1** presents the analytical results of MVGG19 for each of the datasets.

**Table 1**. MVGG19 Results

| Dataset | Accuracy (%) | Precision (%) | Recall (%) | F-1 (%) | AUC (%) |
|---|---|---|---|---|---|
| Casting | 77.62 | 77.68 | - | - | 94.94 |
| Defect | 97.88 | 96.49 | 98.57 | 97.5 | 99.59 |
| Magnetic | 92.67 | 98.49 | - | - | 97.61 |
| Tech | 94.23 | 98.13 | - | - | 99.94 |
| Bridge | 99.02 | 99.5 | 99.36 | 99.43 | 99.83 |
| Solar | 76.78 | 75.63 | 67.02 | 70.93 | 83.36 |

Due to limitations of size, two figures presenting the best and the worst results are illustrated. The rest of the images can be found in the supplementary material. In **Figure 8**, the results for the Casting dataset are presented. In **Figure 9**, the results for the Bridge dataset are presented.

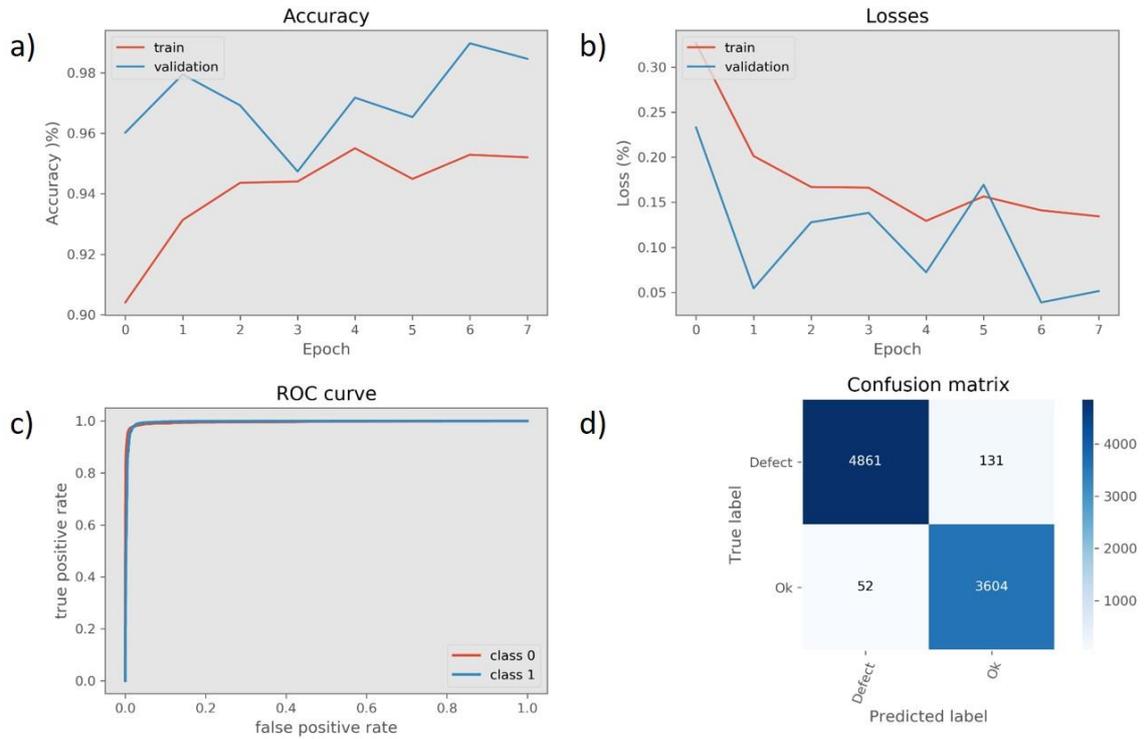

**Figure 8.** Results of the Casting dataset. In a the training and validation accuracy over the epochs of training are presented. In b, the losses. In c, the ROC curves for the two classes, and in d, the confusion matrix.

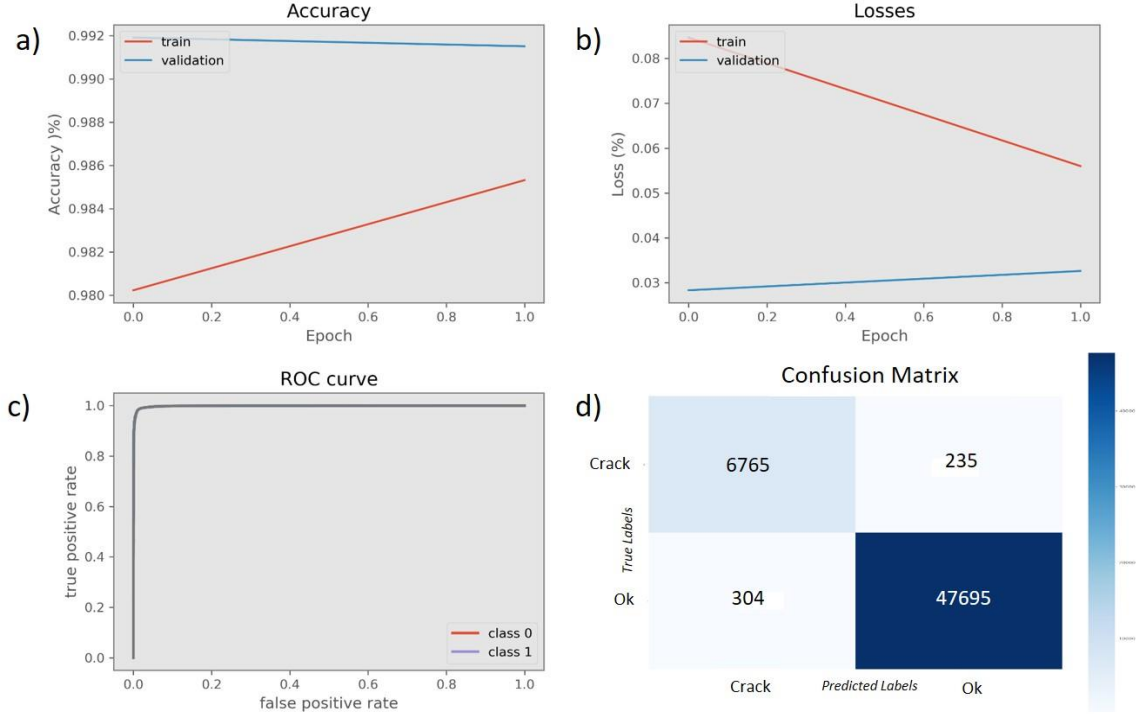

**Figure 9.** Results of the Bridge dataset. In a, the training and validation accuracy over the epochs of training are presented. In b, the losses. In c, the ROC curves for the two classes, and in d, the confusion matrix.

Top-performing scores were observed for most of the datasets, while the confusion matrixes confirm the optimal performance of MVGG19 in industrial object and defect recognition challenges.

The same experiments were performed utilizing the traditional VGG19 sequential structure to evaluate the proposed multipath methodology, retaining the parameters and the hyperparameters as described.

**Table 2**. VGG19 Results. Recall and F-1 score was not recorded for the Defect, Magnetic, and Tech datasets, due to the existence of several classes. Those cases are marked with the "-" symbol in the table.

| Dataset | Accuracy (%) | Precision (%) | Recall (%) | F-1 (%) | AUC (%) |
|---------|--------------|---------------|------------|---------|---------|
| Casting | 87.39 | 78.01 | 97.72 | 87.37 | 97.88 |
| Defect | 70.9 | 66.79 | - | - | 92.65 |
| Magnetic | 77.32 | 70.52 | - | - | 89.16 |
| Tech | 88.29 | 95.31 | - | - | 99.6 |
| Bridge | 98.72 | 99.24 | 99.29 | 99.26 | 99.55 |
| Solar | 73.85 | 74.76 | 58.15 | 64.97 | 80.18 |

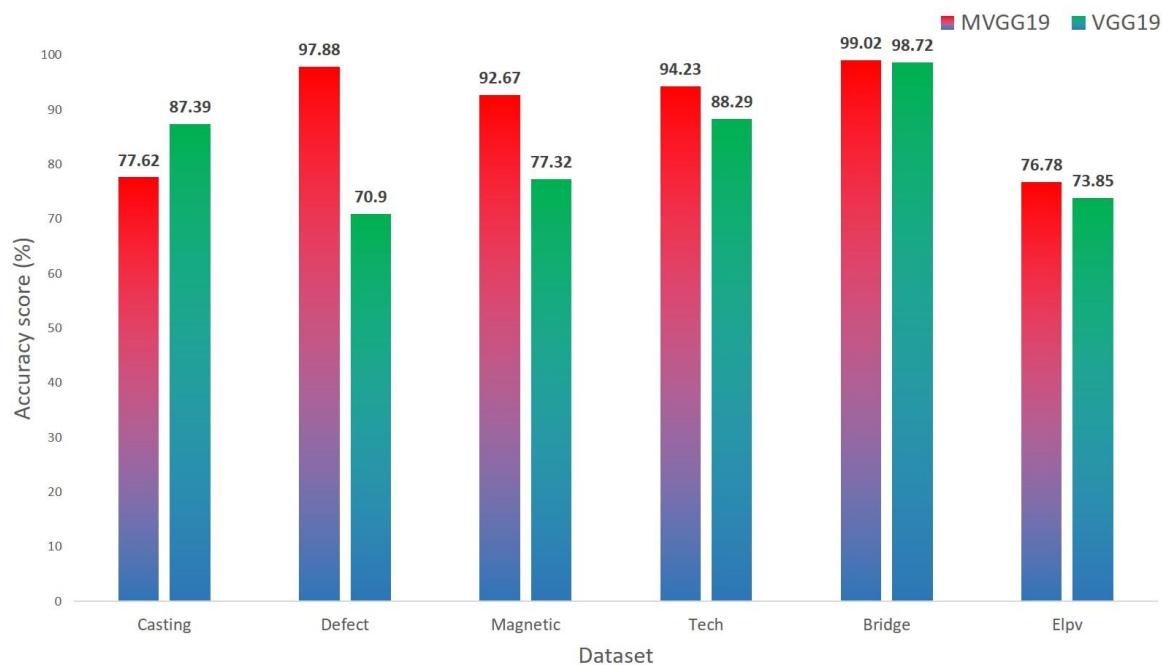

**Figure 10.** Comparison of the obtained overall accuracy of MVGG19 and VGG19. The Elpv bars correspond to the Solar dataset.

The results highlight the superiority of MVGG19 over the traditional VGG19 approach in every object and defect recognition dataset utilized in the present study. The comparison between the two networks in terms of the overall accuracy is provided in **Figure 10**.

**4. Discussion**

MVGG19 was evaluated utilizing six different image datasets, consisting of various classes. Its superiority over the classic VGG19 architecture was demonstrated in five of the six datasets. VGG19 performed quite well, achieving 87.39% accuracy on Casting dataset, 70.9% on the Defect dataset, 77.32% on the Magnetic dataset, 88.29% on the Tech dataset, 98.72% on the Bridge dataset, and 73.85% on the Solar dataset. With MVGG19 the accuracy on the Casting dataset was decreased by 9.77%. The accuracy regarding the Defect dataset was drastically increased by 26.88%. As far as the Magnetic dataset is considered, an improvement of 15.35% was observed. For the Tech dataset, MVGG19 improved the classification accuracy by 5.94%. For the Bridge and the Solar datasets, the accuracy was increased by 0.3% and 2.93% respectively.

The results highlight the effectiveness of the selected feature extraction pipeline. In contrast with the VGG19 architecture, wherein the convolutional layers are stacked in a hierarchical way, MVGG19 allows for staged feature extraction. In this way, feature maps produced by all the convolutional blocks are treated as unique features and not only as derivative features that are simply used by deeper layers to reproduce new ones. The hypothesis that each produced feature map can be significant by itself is confirmed by the results, as MVGG19 succeeds in most of the experimental datasets.

Object detection is an essential asset in the industry that could drastically transform some day-to-day operations that are time consuming and, at the same time, expensive. Finding a specific object through visual inspection is a basic task that is involved in multiple industrial processes like sorting, inventory management, machining, quality management, packaging, etc. Until recently, the quality control part of the manufacturing cycle continues to be a difficult task due to its reliance on human-level visual understanding and adaptation to constantly changing conditions and products. With Artificial Intelligence, most of these complications can be handled. AI can automatically distinguish good parts from faulty parts on an assembly line with incredible speed, allowing enough time to take corrective actions. This is a very useful solution for dynamic environments where product environments are constantly changing and time is valuable to the business. Another aspect for further research is manual sorting which involves high cost of labor and accompanying human errors. Even with robots, the process is not accurate enough and is still prone to a discrepancy. With AI-powered Object Tracking, the objects are classified as per the parameter selected by the manufacturer and statistics of the number of objects is displayed. It significantly reduces the abnormalities in categorization and makes the assembly line more flexible. AI-powered Object Detection can help transform this tedious and manual process into an efficient and automated process while maintaining the same if not better level of accuracy.

Deep Learning networks are a vital part of the algorithms involved in such recognition tasks. For a successful, trustworthy, and global framework to be employed, special-designed models and algorithms are necessary. It is undeniable that not every model is effective in any task. Therefore, the construction of specialized models for industrial image recognition tasks is an interesting and challenging aspect.

## 5. Conclusions

The contributions of this study are two-fold. Firstly, a novel modification proposal for the successful network architecture called VGG is proposed and evaluated for defect object and industrial object recognition tasks. The proposed MVGG19 makes full use of each convolution layer and allows for feature fusion, by concatenating the output features of both top and bottom convolutional layers of the network. In this way, the classification is based on an increased variety of features and results in more precise image analysis. This assumption is confirmed by the results of the experiments. Secondly, since the proposed architecture demonstrated its effectiveness in various industrial image datasets, it is concluded that this architecture can serve as a universal industrial object recognition network and can be employed for related tasks.

**Declarations**

This research did not receive any specific grant from funding agencies in the public, commercial, or not-for-profit sectors.